\documentclass[letterpaper, 10 pt, conference]{ieeeconf}  

\IEEEoverridecommandlockouts                              

\usepackage{bm}
\usepackage{amsfonts} 
\usepackage{mathtools}
\usepackage{amssymb}
\usepackage{amsmath}
\usepackage{graphicx}
\usepackage{caption}
\usepackage{subcaption}
\usepackage{bbm}

\title{\LARGE \bf
Learning robotic cutting from demonstration:\\ Non-holonomic DMPs using the Udwadia-Kalaba method
}

\author{Artūras Straižys$^{1}$, Michael Burke$^{1,2}$ and Subramanian Ramamoorthy$^{1}$
\thanks{$^{1}$School of Informatics, University of Edinburgh}%
\thanks{$^{2}$Department of Electrical and Computer Systems Engineering, Monash University}%
\thanks{For the purpose of open access, the author(s) has applied a Creative Commons Attribution (CC BY) license to any Accepted Manuscript version arising.}
}

\begin{document}

\maketitle
\thispagestyle{empty}
\pagestyle{empty}

\begin{abstract}
Dynamic Movement Primitives (DMPs) offer great versatility for encoding, generating and adapting complex end-effector trajectories. DMPs are also very well suited to learning manipulation skills from human demonstration. However, the reactive nature of DMPs restricts their applicability for tool use and object manipulation tasks involving non-holonomic constraints, such as scalpel cutting or catheter steering. In this work, we extend the Cartesian space DMP formulation by adding a coupling term that enforces a pre-defined set of non-holonomic constraints. We obtain the closed-form expression for the constraint forcing term using the Udwadia-Kalaba method. This approach offers a clean and practical solution for \textit{guaranteed} constraint satisfaction at run-time. Further, the proposed analytical form of the constraint forcing term enables efficient trajectory optimization subject to constraints. We demonstrate the usefulness of this approach by showing how we can learn robotic cutting skills from human demonstration.
\end{abstract}


\section{Introduction}
\begin{figure}[t]
\centering
\includegraphics[width=0.48\textwidth]{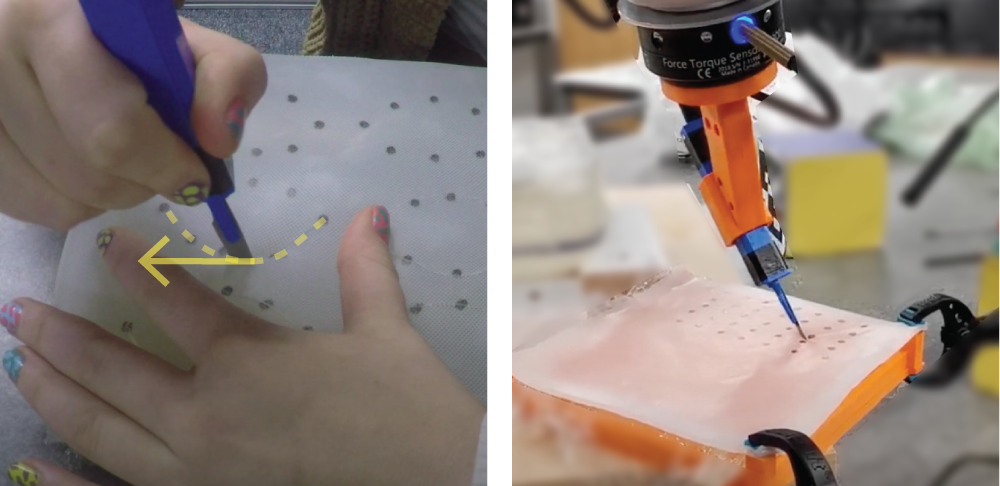}
\caption{Cutting tasks feature non-holonomic constraints that must be considered in automation: a pure lateral movement of the blade is undesired, as it causes unnecessary stress to the material and risks tearing, an unwanted mode of fracture propagation. }
\label{fig:student_cutting}
\end{figure}

Cutting deformable and soft materials is an important skill required in applications ranging from the domestic (e.g. food preparation) to the highly technical (e.g. surgical tasks). A characteristic feature of the cutting task is the prevalence of non-holonomic constraints (Fig. \ref{fig:student_cutting}).
When cutting, it is unnatural and undesirable to move the blade along the lateral direction, as it can induce uncontrolled material tearing. Instead, one aims at neat and predictable material separation \cite{LiuZhihua2022RAiS}. If we were to automate the performance of these tasks by a robot, then that autonomous system must account for these types of constraints, especially when precision plays a crucial role such as in surgery.

Non-holonomic systems attract substantial research interest, especially in the area of mobile robotics \cite{277235,DeLuca1998}. More recently, it has been extensively studied in the context of steerable needles \cite{doi:10.1177/0278364906065388,1570829}, as well as in robotic cutting \cite{zhao2021robotic}, particularly in the tele-operation setting \cite{8968494,7353978,6630766}.
In this paper, we present a novel strategy for \textit{Learning} from Demonstration (LfD) the task of robotic cutting subject to these non-holonomic constraints.
Specifically, we focus on the use of Dynamical Movement Primitives (DMPs) as a representation of the policy. DMPs provide an intuitive way to encode complex kinematic trajectories.
The underlying attractor dynamics of DMPs ensure the system's convergence to a goal state, and offer robustness against uncertainties and perturbations.
However, the reactive nature of DMPs makes it difficult to encode behaviours subject to non-holonomic constraints. In contrast to systems only under holonomic constraints, such as contact constraints, the non-holonomic systems are path-dependent; therefore, a reactive movement adaptation requires global trajectory re-planning for a non-holonomic system to reach the goal state. Besides, it is challenging to capture faithful end-effector orientation with kinesthetic demonstrations. As a result, this further complicates the application of LfD to non-holonomic systems. 

The main contribution of this paper is a scheme to alleviate these issues, demonstrated in the cutting task, wherein we first encode the pose trajectory using two uncoupled DMPs, a position DMP and an orientation DMP. Then, we extend the position DMP with an analytically-derived coupling term that imposes a non-holonomic equality constraint. Finally, we optimize the unconstrained orientation DMP to minimize the constraint coupling term.
While many researchers have proposed various coupling terms for tackling different types of constraints in DMPs \cite{9347805}, e.g. obstacle \cite{hoffmann2009biologically} or surface \cite{han2019trajectory} constraints, in this paper, we focus on the ``rolling-without-slipping'' constraint \cite{murray2017mathematical}, which is relevant in many practical applications such as cutting or catheter steering.
Moreover, we derive a closed-form expression for the constraint coupling term using the Udwadia-Kalaba (UK) theory \cite{udwadia}, for the first time in the DMP literature to the best of our knowledge.

We organize the paper as follows. Firstly, we provide background on DMPs for encoding pose trajectories, and briefly review the Udwadia-Kalaba formalism. Next, we describe our method for incorporating non-holonomic constraints into the DMP framework via an additional coupling term derived from the Udwadia-Kalaba method. Finally, we provide a numerical example and a practical cutting experiment to illustrate the applicability of the proposed approach in an LfD setting.

\section{Background}

\subsection{Dynamic movement primitives}
In the DMP framework \cite{schaal2006dynamic}\cite{Ijspeert}, a movement is represented by a dynamical system that consists of a linear second-order attractor and a nonlinear forcing term:
\begin{equation}
    \tau \ddot{x} = \alpha_x(\beta_x(x_g - x) - \tau\dot{x}) + f_p(s).
\end{equation}
Here, $x$ and $\dot{x}$ are the position and velocity of the system, $x_g$ is the goal position, $\tau$ is a temporal scaling constant, $\alpha_x$ and $\beta_x$ are constants that define the system response. $s$ is a phase variable that acts as a system clock and is defined as follows:
\begin{equation}
    \tau \dot{s} = - \alpha_s s,
\end{equation}
where $\alpha_s$ is a time constant. 

The nonlinear forcing term $f_p$ is responsible for capturing the complexity of the learned trajectory, and is typically fit to demonstration data using the following function approximation:
\begin{equation}
    f_p(s) = \frac{\sum_{i=1}^N \psi_i(s)\theta_i s}{\sum_{i=1}^N \psi_i (s)},
\end{equation}
where $N$ is the number of nonlinear radial basis functions (RBFs) $\psi_i$ and learned parameters $\theta_i$.

In addition to many of its useful properties, such as global asymptotic convergence, spatial and temporal invariance, the above framework enables online adaptation of learned movements via additional coupling terms \cite{5152423}\cite{Pastor}. In this work, we introduce a forcing term that guarantees the compliance with a pre-defined set of non-holonomic constraints.

\subsection{Orientation DMPs}
The original DMP formulation described above is best suited to encoding trajectories in joint angle space or position trajectories in Cartesian space. However, orientations belong to the special orthogonal group SO(3), which is a manifold. Thus, the above formulation is not suitable for use in traditional integration schemes that assume flat Euclidean spaces. Ude et al. in \cite{nemec} proposed a formulation of orientation DMPs that preserves the SO(3) structure, as follows: 
\begin{equation}
    \begin{array}{lcl} \label{orientation_dmp}
    \tau \dot{\bm{w}} = \alpha_x(\beta_x \ \textrm{log}(\bm{R}_g\bm{R}^{T}) - \tau\bm{w}) + \bm{f}_q(s), \\
    \tau \dot{\bm{R}} = \tau [\bm{w}] \bm{R}.
    \end{array}
\end{equation}
where $\dot{\bm{w}} \in \mathbb{R}^3$ is angular acceleration vector, $\bm{w} \in \mathbb{R}^3$ is angular velocity vector, $[\bm{w}] \in \mathbb{R}^{3 \times 3}$ is a skew-symmetric matrix of $\bm{w}$ and $\bm{f}_q(s) \in \mathbb{R}^3$ is nonlinear forcing function for orientation. $\bm{R}$ and $\bm{R}_g \in \textrm{SO(3)}$ are rotation matrices that uniquely specify the current and goal orientations, respectively. 

Assuming constant angular velocity $\bm{w}$, the above differential equation on $\textrm{SO(3)}$ has the following closed-form solution:

\begin{equation}
    \bm{R} = \textrm{exp}([\bm{w}] \Delta t) \bm{R}_0,
\end{equation}
where $\bm{R}_0$ is the initial orientation and $\textrm{exp}(\cdot)$ is the exponential mapping given by the Rodrigues' formula:

\begin{equation} \label{exponential_map}
    \textrm{exp}([\bm{w}]\Delta t) = \bm{I}_3 + \frac{\textrm{sin}(\theta)}{\theta} [\bm{w}] + \frac{1 - \textrm{cos}(\theta)}{\theta^2}[\bm{w}]^2,
\end{equation}

where $\theta = ||\bm{w}||$.

Its inverse, the logarithmic map from eq. (\ref{orientation_dmp}), is defined as the following:

\begin{equation}
\begin{array}{lcl}
    \textrm{log}(\bm{R}) = 
        \begin{dcases}
          [0,0,0]^{T}, & \bm{R} = \bm{I}_3 \\
          \bm{w} = \theta \bm{n}, & \textrm{otherwise.}
        \end{dcases}  \\ \\
        
    \theta = \textrm{arccos}\Big( \frac{\textrm{trace}(\bm{R})-1}{2} \Big), \quad \bm{n} = \frac{1}{2\textrm{sin}(\theta)} \begin{bmatrix} r_{32} - r_{23} \\ r_{13} - r_{31} \\ r_{21} - r_{12} \end{bmatrix}.
    
\end{array}
\end{equation}

\subsection{Constrained dynamics}
In this section, we describe constrained dynamical systems with an explicit form of constraint forces. First, let us consider an unconstrained dynamical system described by configuration space $\mathcal{Q}$, which is an open subset of $\mathbb{R}^n$ with coordinates $\bm{\sigma} = [\sigma_1, ..., \sigma_n]^T$, as follows:

\begin{equation} \label{dynamsys}
    \bm{M}(\bm{\sigma})\ddot{\bm{\sigma}} = \bm{f}_{\textrm{unc}}(\bm{\sigma},\bm{\dot{\sigma}}),
\end{equation}
where $\bm{M} \in \mathbb{R}^{n \times n}$ is the mass inertia matrix, and $\ddot{\bm{\sigma}}$, $\dot{\bm{\sigma}}$ and $\bm{f}_{\textrm{unc}}$ denote generalized unconstrained acceleration, velocity and known unconstrained forcing term, respectively. 

The motion of the above system can be restricted by introducing a set of $k < n$ equality constraints in the following form:
\begin{equation} \label{holonomic}
    \bm{H}(\bm{\sigma}) = 0, 
\end{equation}

where $\bm{H}$ is $k \times n$ constraint matrix. 

The above constraints restrict the configuration space of the unconstrained system to $n-k$ dimensional submanifold and are called \textit{holonomic}. In contrast, the \textit{non-holonomic} constraints restrict the instantaneous motion but not the configuration space of the system. These velocity constraints can be described with the following expression:

\begin{equation} \label{pfaff}
    \frac{d}{dt} \bm{H}(\bm{\sigma}) =  \frac{\partial \bm{H}(\bm{\sigma})}{\partial \bm{\sigma}} \dot{\bm{\sigma}} = 0
\end{equation}

The dynamics of the constrained system can be viewed as a system subjected to constraint forces $\bm{f}_{\textrm{con}}$, as follows \cite{murray2017mathematical}:

\begin{equation}
\begin{split}
    \bm{M}(\bm{\sigma})\ddot{\bm{\sigma}} &= \bm{f}_{\textrm{unc}}(\bm{\sigma},\bm{\dot{\sigma}}) + \bm{f}_{\textrm{con}} \\
    &= \bm{f}_{\textrm{unc}}(\bm{\sigma},\bm{\dot{\sigma}}) + \frac{\partial \bm{H}(\bm{\sigma})}{\partial \bm{\sigma}}^T \bm{\lambda}
\end{split}
\end{equation}
where $\bm{\lambda}$ is a vector of Lagrange multipliers.

The computation of the Lagrange multipliers $\bm{\lambda}$ is often a difficult task. With an exception of the simplest cases, the Lagrange multipliers need to be solved for numerically. Moreover, the Lagrange multiplier is generally not unique for given constraint force \cite{Zhao2018}. In the next section, we present a description of the Udwadia-Kalaba method for deriving an analytical expression of constraint forces $\bm{f}_{\textrm{con}}$, which greatly simplifies the task. 

\subsection{Udwadia-Kalaba method}\label{UK}

Udwadia and Kalaba presented a novel approach for deriving the analytical expression of motion for dynamical systems under the holonomic and non-holonomic equality constraints \cite{udwadia}.
In comparison to other methods that rely on Lagrange multipliers or other auxiliary variables, the UK method provides an explicit expression of the constraint forces based on the generalized coordinates alone. 

Consider an unconstrained dynamical system (\ref{dynamsys}) with identity mass inertia matrix, i.e. $\bm{M} = \bm{I}_n$. Suppose that this system is subjected to a set of $m$ equality constraints of the following form:
\begin{equation} \label{pfaffian}
    \bm{A}(\bm{\sigma},\bm{\dot{\sigma}}) = \bm{0},
\end{equation}
where $\bm{A} \in \mathbb{R}^{m \times n}$ is constraint matrix, which could include holonomic or non-holonomic constraint equations, or a combination of both.

A key step in the UK approach is the transformation of the above constraint equation into a second order form, which is achieved by taking the time derivative of the constraint equation twice (under the assumption that it is sufficiently smooth):

\begin{equation} \label{second-order}
    \bm{A}(\bm{\sigma},\bm{\dot{\sigma}}) \ddot{\bm{\sigma}} = \bm{b}(\bm{\sigma},\bm{\dot{\sigma}}),
\end{equation}
where $\bm{b} \in \mathbb{R}^m$ is known generalized velocity vector. 

Finally, the dynamical system under constraint (\ref{second-order}) can be now expressed as:
\begin{equation}\label{constrained-dynamics}
    \ddot{\bm{\sigma}} = \bm{f}_{\textrm{unc}}(\bm{\sigma},\bm{\dot{\sigma}}) + \bm{f}_{\textrm{con}}(\bm{\sigma},\bm{\dot{\sigma}}, \bm{f}_{\textrm{unc}}),
\end{equation}
where $\bm{f}_{\textrm{con}} \in \mathbb{R}^n$ denotes the constraint forcing term, whose closed-form is provided by the following UK equation:

\begin{equation}\label{udwadia-kalaba-eq}
    \bm{f}_{\textrm{con}}(\bm{\sigma},\bm{\dot{\sigma}}, \bm{f}_{\textrm{unc}}) = \bm{A}^+(\bm{b} - \bm{A}\bm{f}_{\textrm{unc}}),
\end{equation}
Here, superscript $^+$ indicates the Moore-Penrose pseudoinverse.

It should be noted, that the UK formulation of constrained dynamics assumes that the constraint is satisfied by the initial condition.

\section{Non-holonomic DMPs}
\subsection{UK-based coupling term for enforcing non-holonomic constraint}
Consider a non-holonomic constraint of the following form:

\begin{equation} \label{example-constraint}
    \bm{c}^T \dot{\bm{p}} = 0,
\end{equation}
where $\bm{c}$ and $\dot{\bm{p}} \in \mathbb{R}^3$ are the constraint and instantaneous velocity vectors, respectively, both expressed in the world frame. 

We can turn the above constraint equation into second-order form by differentiating it with respect to time:

\begin{equation}
    \dot{\bm{c}}^T \dot{\bm{p}} + \bm{c}^T \ddot{\bm{p}} = 0,
\end{equation}
where $\ddot{\bm{p}} \in \mathbb{R}^3$ is vector of instantaneous unconstrained acceleration.

By comparison with (\ref{second-order}), we can identify the generalized velocity vector as $\bm{b} = - \dot{\bm{c}}^T \dot{\bm{p}}$.  Then, we can apply UK formula (\ref{udwadia-kalaba-eq}) to express the non-holonomic constraint forces as follows:

\begin{equation} \label{dmp-constraint-force}
    \bm{f}_{\textrm{con}} = \bm{c}^+ (\bm{b} - \bm{c}^T \ddot{\bm{p}})
\end{equation}

The above analytical expression for constraint forces $\bm{f}_{\textrm{con}}$ can be incorporated as an additional DMP coupling term. The encoding of the end-effector's pose under constraint (\ref{example-constraint}) can be achieved by composition of constrained DMP for position and an unconstrained DMP for orientation, as follows:

\begin{equation} \label{pose-dmp}
    \begin{array}{lcl}
      \tau \ddot{\bm{p}}_{\textrm{con}} = \underbrace{\alpha_x(\beta_x(\bm{p}_g - \bm{p}) - \tau\dot{\bm{p}}) + \bm{f}_p(s)}_{\bm{\ddot{p}}} \ + \ \bm{f}_{\textrm{con}}, \\
      \tau \dot{\bm{w}} = \alpha_x(\beta_x \ \textrm{log}(\bm{R}_g\bm{R}^{T}) - \tau\bm{w}) + \bm{f}_q(s),
    \end{array}
\end{equation}
where $\ddot{\bm{p}}_{\textrm{con}} \in \mathbb{R}^3$ is constrained acceleration vector, $\bm{p}$ and $\bm{p}_g \in \mathbb{R}^3$ are current and goal position vectors, respectively.

The $\bm{f}_{\textrm{con}}$ term in the above formulation (\ref{pose-dmp}) guarantees the satisfaction of the constraint eq. (\ref{example-constraint}) by modifying system's acceleration at run-time. In addition, the explicit expression for $\bm{f}_{\textrm{con}}$ enables efficient trajectory optimization for $\bm{f}_{\textrm{con}} = \bm{0}$, such that

\begin{itemize}
    \item the constraint eq. (\ref{example-constraint}) is satisfied,
    \item deviations from the demonstrated position trajectory are minimized. 
\end{itemize}

In the following section, we show how the optimization of the orientation DMPs achieves the above objectives in learning autonomous cutting skills from a demonstration. 


\begin{figure*}[t]
\centering
\includegraphics[width=\textwidth]{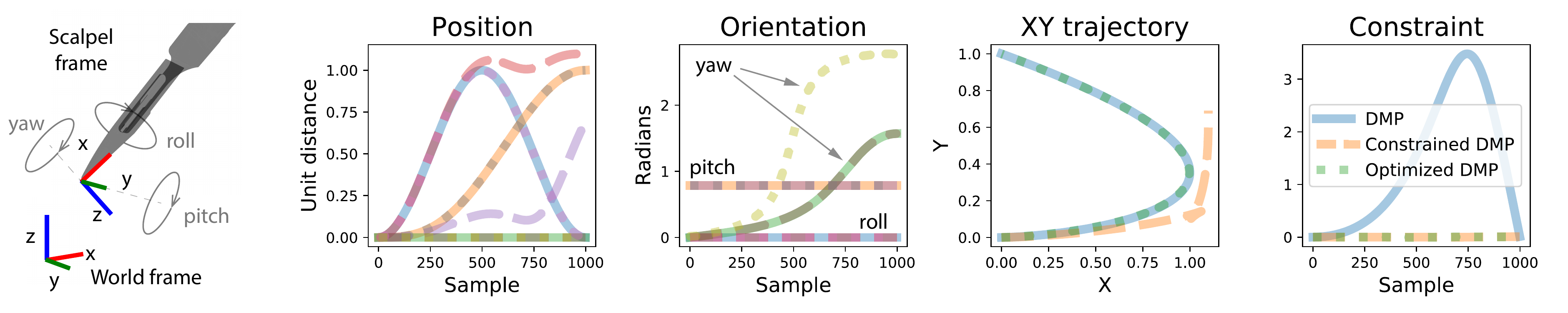}
\caption{(Left) Scalpel frame. (Other) Regular DMP (solid lines) imitates the demonstrated cutting trajectory that closely follows the desired curved contour on the $XY$ plane, but violates the constraint. The DMP constrained by $\bm{f}_{\textrm{con}}$ term (dashed lines) satisfies the constraint and closely follows (unconstrained) orientation trajectory, but deviates from the (constrained) $x$ and $y$ trajectories. The DMP with optimized orientation (dotted lines) closely follows the desired position and orientation trajectories, and satisfies the constraint.}
\label{fig:numerical_example_plots}
\end{figure*}

\begin{figure}
\centering
\includegraphics[width=0.48\textwidth]{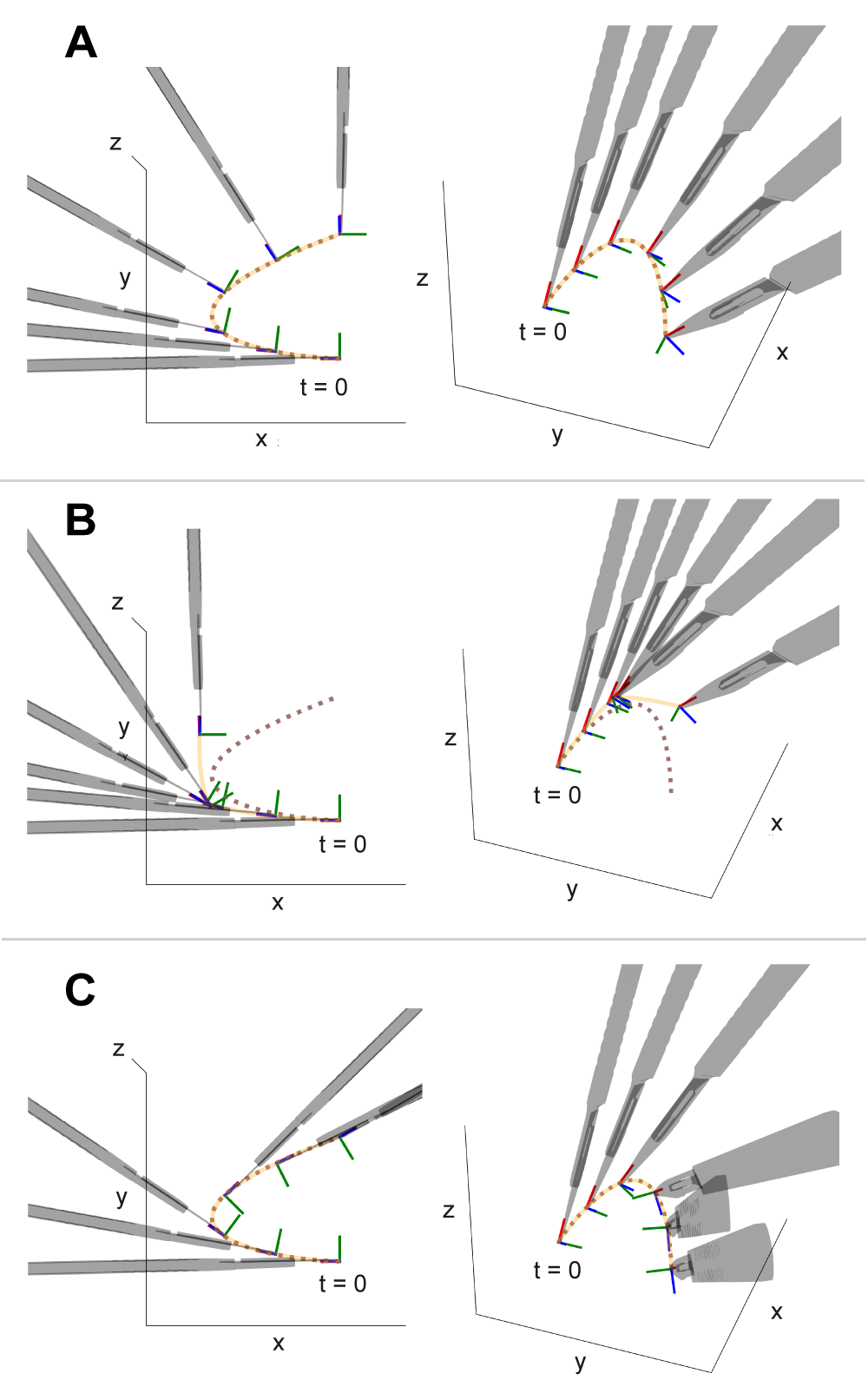}
\caption{Snapshots of scalpel trajectories. \textbf{A} Demonstration that violates the non-holonomic constraint. Note, that blade moves in the prohibited lateral direction ($\hat{\bm{y}}_b$, green) towards the end of task execution. \textbf{B} DMP constrained by the coupling term $\bm{f}_{\textrm{con}}$. \textbf{C} DMP with optimized orientation for $\bm{f}_{\textrm{con}} = \bm{0}$. Note, that the optimized DMP closely follows the desired path on $XY$ plane (marked by dotted line) and $\hat{\bm{y}}_b$ vector is normal to a contour tangent.}
\label{fig:numerical_example_snapshots}
\end{figure}

\subsection{Learning cutting skills from demonstration}
Consider learning scalpel cutting skills from demonstration using DMPs under the non-holonomic constraint, as described in the previous section. 
Fig. \ref{fig:numerical_example_plots} (left) shows the local body-frame of the scalpel with respect to the world frame. Let's denote the coordinates of body-fixed frame as $(\hat{\bm{x}}_b,\hat{\bm{y}}_b,\hat{\bm{z}}_b)$, where $\hat{\bm{x}}_b = [1,0,0]^T$, $\hat{\bm{y}}_b=[0,1,0]^T$ and $\hat{\bm{z}}_b = [0,0,1]^T$. Let the rotation matrix describing the orientation of the body-fixed frame with respect to the world frame be denoted as $\bm{R}$. 
In this task, we want to prohibit the lateral motion of the scalpel, thus we define the constraint vector from eq. (\ref{example-constraint}) as $\bm{c} = \bm{R}\hat{\bm{y}}_b$. Corresponding $\bm{b}$ vector from eq. (\ref{dmp-constraint-force}) is therefore equal to $\bm{b} = - (\dot{\bm{R}}\hat{\bm{y}}_b) ^T \dot{\bm{p}}$. Recall that $\dot{\bm{R}} = [\bm{w}] \bm{R}$, thus the constraint force vector $\bm{f}_\textrm{con}$ is a function of angular velocity $\bm{w}$. It is, therefore, possible to optimize the rotation for $\bm{f}_{\textrm{con}} = \bm{0}$, as described in the previous section. 

Solving for $\bm{f}_{\textrm{con}}(\bm{w}) = \bm{0}$ is a nonlinear problem, which can be approached numerically. The $\bm{w}$ solution can be found by minimizing the following loss function $\mathcal{L}(\bm{w})$:
\begin{equation}
    \mathcal{L}(\bm{w}) = \lVert\bm{f}_{\textrm{con}} (\bm{w})\rVert_2 \ + \ \lVert\bm{R} - \bm{R}_{\textrm{opt}}(\bm{w})\rVert_F
\end{equation}
where $\bm{R}_{\textrm{opt}}(\bm{w})$ is optimized rotation, $\lVert \ \cdot \ \rVert_2$ is the Euclidean norm and $\lVert \ \cdot \ \rVert_F$ is the Frobenius norm.

The first term on the loss $\mathcal{L}$ penalizes the constraint forces $\bm{f}_{\textrm{con}}$ and the second term minimizes the distance between the original and optimized rotations. In the next section, we demonstrate the application of the described method using a numerical example followed by a practical experiment.

\section{Experiments}
\subsection{Numerical example}

Here we provide a concise illustration of the proposed non-holonomic DMP formulation used to learn a cutting skill from demonstration. We assume that cutting is required along a desired curved contour on the $XY$ plane. As before, we represent the cutting task with a kinematics model that prohibits lateral motion of the blade (the coordinate system of the scalpel is shown in Fig. \ref{fig:numerical_example_plots}, left). In this example, we assume that the demonstration successfully traces the desired cutting contour, but violates the constraint, as illustrated in Fig. \ref{fig:numerical_example_snapshots}A. 

Fig. \ref{fig:numerical_example_plots} compares the resulting trajectories of the regular DMP (solid lines), the DMP constrained by constraint coupling term $\bm{f}_{\textrm{con}}$ (dashed lines) and DMP with an optimized orientation for $\bm{f}_{\textrm{con}} = \bm{0}$ (dotted lines). As expected, the regular DMP closely follows the demonstration for both position and orientation components. Since the demonstrated pose trajectory violates the constraint, so does the regular DMP. The constrained DMP closely imitates the demonstrated unconstrained orientation trajectory. However, the system deviates from the desired position trajectory as the result of acting constraint forces $\bm{f}_{\textrm{con}}$. Finally, the DMP with optimized angular velocity $\bm{w}$ shows the adjustments in orientation trajectory (note the yaw angle) for maintaining zero constraint forces $\bm{f}_{\textrm{con}}$. As the result, the system follows the desired position trajectory without violating the constraint. Fig. \ref{fig:numerical_example_snapshots} shows the corresponding poses of the scalpel in 3D. 

In this example, the demonstration position trajectories were $x(t) = \textrm{sin}^2(\pi t)$, $y(t) = \textrm{sin}^3(\frac{1}{2} \pi t)$ and $z(t) = 0$ for all $t$. The orientation trajectories were generated with $\alpha(t) = 0$ for roll angle, $\beta(t) = \frac{\pi}{4}$ for pitch angle, and $\gamma(t) = \textrm{arctan2}(x(t),y(t))$ for yaw angle. The number of radial basis functions in DMPs was set to 100, with $\tau=1$, $\alpha_x = 25$, $\beta_x = 6.25$ and $\alpha_s = 1$. The DMPs were integrated using the Implicit Euler method, with time step $\Delta t = 0.001$s. For optimization, we used implementation of BFGS algorithm in SciPy \cite{2020SciPy-NMeth}. The visualization of the scalpel trajectories in Fig. \ref{fig:numerical_example_snapshots} was implemented using pytransform3d library \cite{Fabisch2019}.

\subsection{Learning elliptical excision from demonstration under non-holonomic constraints}

\begin{figure}[t]
\centering
\includegraphics[width=0.45\textwidth]{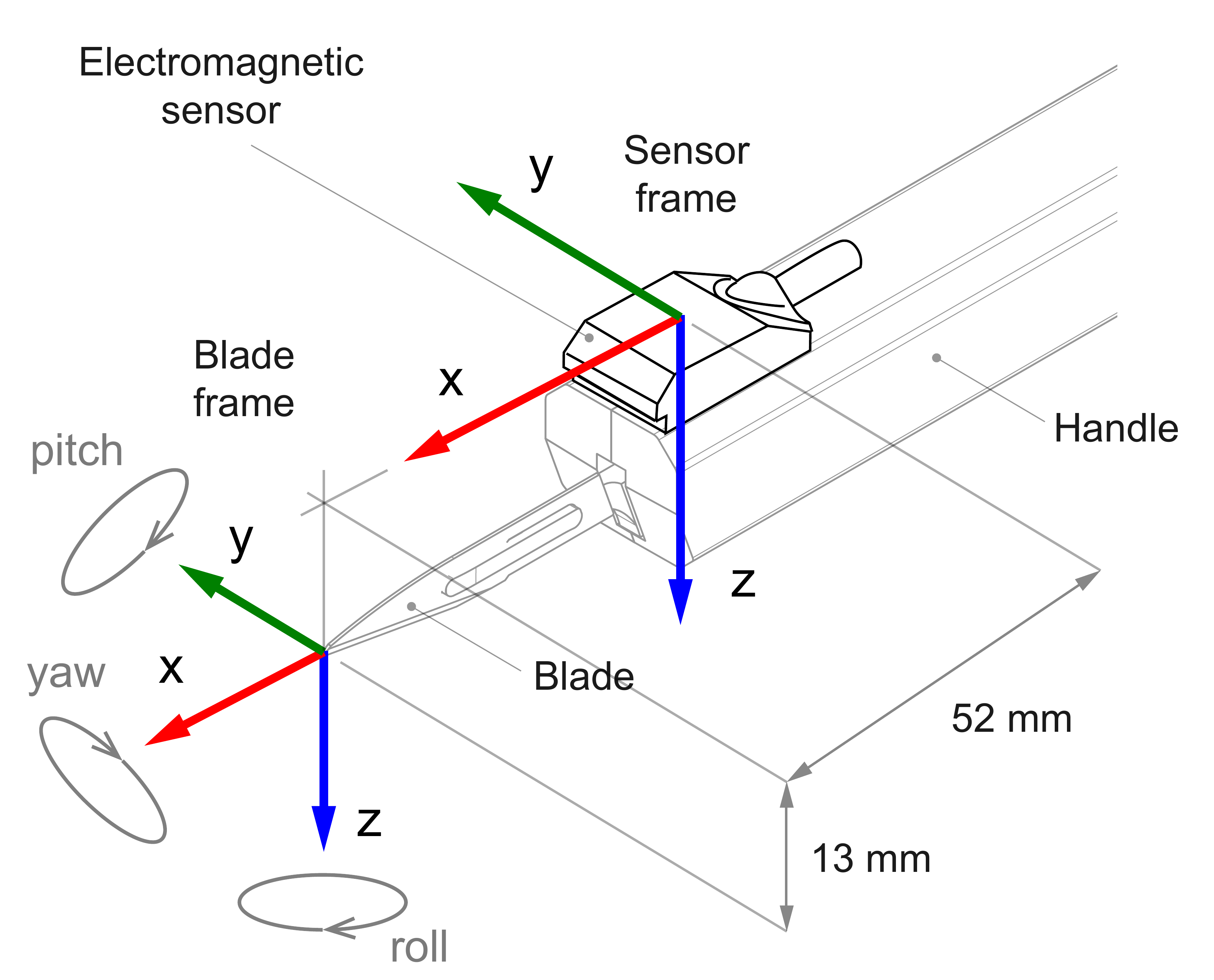}
\caption{Schematic diagram of the cutting tool and sensor configuration, and the coordinate frames used in the experiment.}
\label{fig:sensor_arrangement}
\end{figure}

In this experiment, we apply the proposed approach for learning non-holonomic DMPs from demonstration to the case of elliptical tissue excision (Fig. \ref{fig:student_cutting}, left).
The experimental setup consists of a tissue phantom and a cutting tool with a mounted 6DOF electromagnetic motion tracking sensor (Polhemus VIPER$^{\textrm{TM}}$ FT-Flatsided), as shown in Fig. \ref{fig:sensor_arrangement}. 

During the experiment, the demonstrator performed a series of eight elliptical excisions on the phantom. 
The pose trajectory of the sensor was captured at a fixed rate of 120 Hz and converted to a corresponding pose trajectory of the blade, as follows. We define the blade frame with respect to the sensor frame using following homogeneous transformation:

\begin{figure}[t]
\centering
\includegraphics[width=0.48\textwidth]{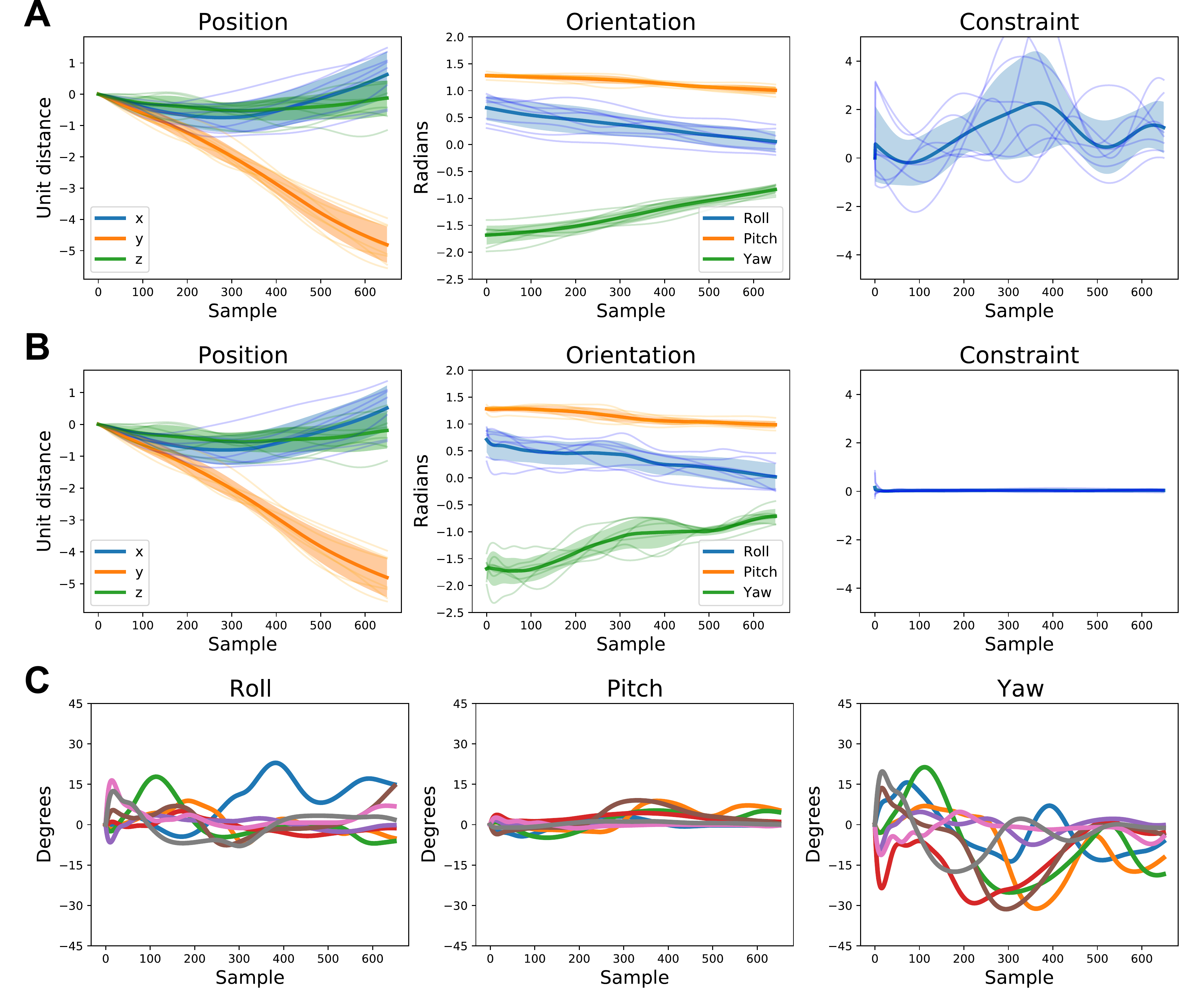}
\caption{\textbf{A} Mean and standard deviation ($N=8$) of blade trajectories in the elliptical excision task (individual trajectories are shown as thin semi-transparent lines). \textbf{B} Trained non-honomic DMPs with orientation optimized for zero $\bm{f}_{\textrm{con}}$. \textbf{C} Adjustments to orientation trajectories of the blade. Note: Roll, pitch and yaw angles follow the extrinsic $xyz$ Euler convention.}
\label{fig:elliptical_optimization}
\end{figure}

\begin{equation}
    \bm{T}_{sb} = \begin{bmatrix} \bm{R}_{sb} & \bm{t}_{sb} \\ \bm{0}_{1 \times 3} & 1 \end{bmatrix} = \begin{bmatrix} 1 & 0 & 0 & 5.2 \\ 0 & 1 & 0 & 0 \\ 0 & 0 & 1 & 1.3 \\ 0 & 0 & 0 & 1 \end{bmatrix},
\end{equation}
where $\bm{R}_{sb} \in \textrm{SO(3)}$ is the rotation matrix representing the orientation of the blade frame in the sensor reference frame, and vector $\bm{t}_{sb} \in R^3$ is the position of the blade frame with respect to the sensor frame, expressed in cm (see Fig. \ref{fig:sensor_arrangement}). The pose trajectory of the blade with respect to the world frame is then simply $\bm{T}_{wb}(t) = \bm{T}_{ws}(t) \bm{T}_{sb}$, where $\bm{T}_{ws}(t)$ is the time-varying homogeneous transformation representing the original pose trajectory of the sensor.

Eight segments of $\bm{T}_{wb}(t)$ trajectory corresponding to each elliptical excision were manually extracted. For each segment of the position trajectory, we subtracted the initial positions such that resultant $XYZ$ trajectories start at the origin (i.e. relative to the initial positions of the blade). Finally, the time series were filtered using a 3rd order Butterworth low-pass filter with 4.8 Hz cut-off frequency. 

As discussed in section \ref{UK}, the UK method requires the initial condition to satisfy the constraint. We enforced this condition by projecting blade's initial velocity vector $\bm{\dot{p}}_0$ onto the $XZ$ plane of the blade frame, as follows:

\begin{equation}
    \bm{\dot{p}}_{0}' = \bm{P}(\bm{P}^T \bm{P})^{-1} \bm{P}^T \bm{\dot{p}}_{0},
\end{equation}
where $\bm{\dot{p}}_{0}'$ is the orthogonal projection of $\bm{\dot{p}}_0$ onto the plane defined by matrix $\bm{P} = [\bm{R}_{wb}\hat{\bm{x}}_b,\bm{R}_{wb}\hat{\bm{z}}_b]$. Here, $\bm{R}_{wb}$ is the rotation matrix representing blade's orientation with respect to the world frame, $\hat{\bm{x}}_b$ and $\hat{\bm{z}}_b$ are basis vectors of the blade frame, respectively.

Finally, the obtained pose trajectories of the blade were used as demonstrations to train the proposed non-holonomic DMPs with optimized orientation (the DMP parameters, i.e. number of RBFs, $\tau$, $\alpha_x$, $\beta_x$ and $\alpha_s$, matched those in the numerical example above).
Fig. \ref{fig:elliptical_optimization}A shows the distribution (mean and $\pm$1 standard deviation) of demonstrations. The evaluated constraint eq. (\ref{example-constraint}) shows that, on average, the constraint violation peaks in the middle of task execution. Partially, this can be explained by the velocity variable $\dot{\bm{p}}$, which reflects the natural acceleration-deceleration motion profile. Alternatively, this can be related to an awkward wrist configuration as the scalpel passes through the peak of the parabola (Fig. \ref{fig:student_cutting}). Interestingly, the constraint curve highlights the systematic violation of the non-holonomic constraint, predominantly along the positive $\hat{\bm{y}}_b$ vector of the blade frame.

Fig. \ref{fig:elliptical_optimization}B shows the distribution of trained non-holonomic DMPs with optimized orientation. As expected, the position trajectories of the blade match the original demonstrations. However, the orientation trajectories show noticeable adjustments, in particular, to roll and yaw angles of rotation. Most importantly, the optimized non-holonomic DMPs satisfy the constraint throughout the entire task execution. Fig. \ref{fig:elliptical_optimization}C shows the individual adjustments to roll, pitch and yaw rotations of the demonstrated pose trajectories. With an exception to an outlier (blue line), most orientation adjustments, as expected, are made to yaw rotation (reaching 30 degrees of angle correction). 

\section{Conclusions}
In this paper, we presented a novel approach to constrain DMPs through an additional coupling term deived from an analytical solution derived using the Udwadia-Kalaba method.
This approach enables the incorporation of a wide range of equality constraints and their combination, such as holonomic, nonholonomic, scleronomic, rheonomic, catastatic and acatastatic, among others \cite{udwadia}. 
The analytical expression of the constraint forcing term allows evaluation of the constraint forces at run-time.
Our experiments demonstrate the efficacy of the proposed method in learning robotic cutting skills from demonstrations under non-holonomic constraints.

The proposed approach does have several limitations, addressing which is a focus of our current and future work. Firstly, the described Udwadia-Kalaba method for the explicit expression of constraint forces applies to equality constraints only. Therefore, it excludes a highly relevant class of constraints, such as joint limits or collision avoidance. Secondly, the UK approach assumes that the constraint is satisfied by the initial condition already (which need not be the case when the task is initialised arbitrarily in practical applications). Finally, the described approach relies on run-time optimization to ensure that the constrained DMP follows the demonstration - this may need further computational treatment for efficient embedded and real-time implementation.

\bibliographystyle{IEEEtran}
\bibliography{references.bib}

\end{document}